# Can Synthetic Data be Fair and Private? A Comparative Study of Synthetic Data Generation and Fairness Algorithms

QINYI LIU, Centre for the Science of Learning & Technology (SLATE), University of Bergen, Norway

OSCAR DEHO, University of South Australia, Australia

FARHAD VADIEE, Centre for the Science of Learning & Technology (SLATE), University of Bergen, Norway

MOHAMMAD KHALIL, Centre for the Science of Learning & Technology (SLATE), University of Bergen, Norway

SRECKO JOKSIMOVIC, University of South Australia, Australia

GEORGE SIEMENS, University of South Australia, Australia

The increasing use of machine learning in learning analytics (LA) has raised significant concerns around algorithmic fairness and privacy. Synthetic data has emerged as a dual-purpose tool, enhancing privacy and improving fairness in LA models. However, prior research suggests an inverse relationship between fairness and privacy, making it challenging to optimize both. This study investigates which synthetic data generators can best balance privacy and fairness, and whether pre-processing fairness algorithms, typically applied to real datasets, are effective on synthetic data. Our results highlight that the DEbiasing CAusal Fairness (DECAF) algorithm achieves the best balance between privacy and fairness. However, DECAF suffers in utility, as reflected in its predictive accuracy. Notably, we found that applying pre-processing fairness algorithms to synthetic data improves fairness even more than when applied to real data. These findings suggest that combining synthetic data generation with fairness pre-processing offers a promising approach to creating fairer LA models.

## 1 Introduction

Machine learning (ML) models have been widely applied in the Learning Analytics (LA) field for various predictive tasks [16, 25]. The widespread use of ML models in LA brings numerous benefits, such as improving students' learning experiences by generating personalized learning paths through the analysis of student behavior [49]. Additionally, ML provides strategic support for optimizing teaching strategies by analyzing students' responses to courses [49]. However, applying ML in LA also raises privacy concerns. Many types of student data, such as mental health data, are considered personal information and must be de-identified according to data protection laws [37]. In this context, many privacy-preserving methods are used in conjunction with ML to meet legal privacy requirements. Among them,

Authors' Contact Information: Qinyi Liu, Centre for the Science of Learning & Technology (SLATE), University of Bergen, Bergen, Norway; Oscar Deho, University of South Australia, Adelaide, Australia; Farhad Vadiee, Centre for the Science of Learning & Technology (SLATE), University of Bergen, Bergen, Norway; Mohammad Khalil, Centre for the Science of Learning & Technology (SLATE), University of Bergen, Bergen, Norway; Srecko Joksimovic, University of South Australia, Adelaide, Australia; George Siemens, University of South Australia, Adelaide, Australia.

Synthetic Data Generators (SDGs) are considered a promising method [28]. SDGs protect privacy by allowing the sharing and publication of synthetic data instead of real data containing personal information.

In addition to being used as a privacy-preserving method, SDGs can also improve algorithmic fairness. ML models that are used in LA may have different impacts on minority groups for various reasons, including class imbalance in the underlying training datasets [42] and smaller sample sizes for minority groups [9]. These effects often amplify existing societal biases against minority groups under the guise of automated fairness [10]. Creating balanced synthetic datasets using SDGs to enhance classification training has been shown to help mitigate the disparate impacts caused by minority group imbalances [1, 3].

It can be seen that SDGs can function both as contributing to fairness improvements and privacy-preserving methods. However, the impact of SDGs on fairness and privacy is not entirely positively correlated. SDGs have been proven to enhance either privacy or fairness, and there are dedicated SDGs designed specifically for privacy-preserving purposes (e.g., [29]), as well as SDGs focused on generating fair synthetic data (e.g., [51]). Some studies, meanwhile, indicate that it is challenging for a single SDG to simultaneously improve both privacy and fairness [46]. Specifically, if an SDG performs well in terms of privacy, its improvement in fairness tends to be more limited, and vice versa [46]. Through our literature review, we find that previous research has primarily focused on studying which privacy-preserving SDG balanced well in terms of privacy and utility [45] or focused on studying which SDGs balance well in terms of utility and fairness [45]. There is a lack of studies examining the trade-offs between privacy and fairness across a broader range of SDGs, including privacy-preserving SDGs, fairness-focused SDGs, and general SDGs. This paper aims to fill this gap. Moreover, considering the recent advancements in Large Language Models (LLMs) within the synthetic tabular data [8], this paper also includes an LLM-based SDG in the evaluation. In addition to studying the balance between fairness and privacy in SDGs, we also observed that prior research on improving fairness in synthetic data has predominantly applied fairness constraints during the data generation process [46, 51]. Few studies have explored the use of pre-processing fairness algorithms with SDGs. But pre-processing fairness algorithms have demonstrated good performance on real data [15, 43, 53], and investigating whether they perform similarly on synthetic data could help advance fairness in SDGs. More formally, this paper aims to investigate the following research questions (RQs)

**RQ1**. *Which synthetic data generator performs best in terms of balancing both privacy and fairness?*

**RQ2**. *How can we improve the fairness of synthetic datasets with pre-processing fairness algorithms?*

To answer RQ1, we generate synthetic datasets from 3 real-world datasets using 5 SDGs and evaluate them across 4 widely used privacy metrics. We go on to train 4 popularly used ML models on the synthetic datasets and evaluate the fairness of models using 3 fairness metrics. We finally perform the privacy *vs.* fairness analysis. To answer RQ2, we do the following: (1) we *debias* the synthetic datasets that were generated in RQ1 by applying 4 pre-processing fairness algorithms, (2) we train 4 ML models on the debiased data, and (3) we investigate the improvement (or lack thereof) in fairness in terms of the 3 fairness metrics. Furthermore, we evaluate the utility of the models for both RQs by examining their predictive accuracy. By answering these questions, we make the following important contributions to privacy and fairness in LA research:

- We perform an extensive and rigorous study that highlights the triangular relationship between privacy, fairness and utility. Specifically, we show that privacy and fairness might have a direct relationship as pair and joint inverse relationship with utility (i.e., predictive accuracy).
- We show that the combination of the synthetic datasets with fairness algorithms results in fairer predictions as compared to the combination of real-world data and fairness algorithms.
- We provide important policy implications and useful recommendations for stakeholders and practitioners based on our findings.

The rest of this paper is organized as follows. In section 2, we provide brief background information about relevant related works. We discuss our datasets, techniques and metrics, and experimental details in section 3. We provide the results of our experiments in Section 4, and finally, we discuss the implications of our findings and conclude the paper in Section 5.

## 2 Related work

### 2.1 Privacy and Fairness in LA

Privacy and fairness have been a central topic of discussion within the community since the inception of LA [17, 31]. This section will explain the definitions of privacy and fairness used in this paper, as well as the current status of technologies to improve privacy and fairness in the LA field.

*2.1.1 Privacy in LA.* In this paper, we adopt the machine learning perspective on privacy, with a focus on individuals' consent for data collection and the prevention of harm through data sharing, in line with the concerns raised by Jordon et al. [28]. As highlighted in the field of LA, the increasing complexity of global data protection laws, such as the EU's GDPR, requires strict anonymization of personal and sensitive data before sharing. However, despite efforts to protect data, traditional anonymization methods often fall short under adversarial attacks, as demonstrated by [56], where anonymized student data was re-identified. This underscores the growing challenges faced in the LA field, where personal data must be handled with precision.

Amid these concerns, advanced privacy-preserving techniques, including SDG, are gaining traction within LA research. Liu et al. [39] demonstrated the effectiveness of various SDG techniques in improving privacy protection while preserving data utility across different dataset sizes. Zhan et al. [59] also emphasized the superiority of differentially private SDGs compared to traditional privacy methods. Although the application of synthetic data as a privacy-preserving measure in LA is relatively new, it shows considerable potential, mirroring its success in other sectors like healthcare and finance [28]. This marks a significant shift towards embracing synthetic data to safeguard personal information without compromising the utility needed for meaningful analytics.

*2.1.2 Fairness in LA.* In recent years, the issue of algorithmic fairness has increasingly captured the focus of research in LA [14, 15, 23, 34, 58]. Several metrics—including inter alia, statistical parity difference, equal opportunity, Absolute Between Receiver Operating Characteristic curve Area (ABROCA)—have been defined to operationalize fairness in LA and the broader ML community [34]. Correspondingly, the research community has designed many so-called fairness algorithms that aim to satisfy one or more fairness metrics to mitigate unfairness in predictive models [34, 43]. Fairness algorithms ensure fairness by either removing unfairness from the training data (i.e., pre-processing), adding some fairness constraint to the objective function of the ML model (i.e., in-processing), or removing unfairness from previous predictions made by a model (i.e., post-procesing). While pre-, in-, and post-processing algorithms each tackle different segments of the ML pipeline, there is general consensus that the key source of algorithmic unfairness is the dataset on which predictive models are trained [34, 43]. This is not unexpected given that most real-world data are laden with various degrees of historical biases [4, 34, 58]. To address the issue of historical biases in real-world data, the generation of synthetic data has become a viable option.

Synthetic data has shown significant potential in enhancing algorithmic fairness. For example, many studies have demonstrated that using balanced synthetic datasets based on generative adversarial networks (GANs) to improve classification training can help reduce the biases caused by imbalances in minority groups [1, 3, 41]. Panagiotou et al. [44] further argue that, unlike traditional sampling techniques used to mitigate imbalance, synthetic data provides a possible solution for addressing both class and group imbalances. Their experiments on four datasets of varying sizes support this claim. In the LA field, while less common than in other domains, recent years have seen a few innovative efforts to use synthetic data to address fairness issues. For example, Jiang et al. [27] use synthetic data to generate unfair benchmark datasets, avoiding the need for actual data collection that may raise ethical concerns (such as sensitive data involving minority student groups).

*2.1.3 The Relationship between Fairness and Privacy.* Synthetic data has emerged as an effective approach for enhancing both privacy and fairness in algorithms. However, previous studies indicate that synthetic data often struggles to balance these two goals [41]. This challenge arises because achieving stronger privacy protection, particularly through the use of differential privacy, can undermine fairness. It is widely recognized that differential privacy can compromise the fairness of synthetic data, leading to increased research focused on evaluating the fairness of differentially private synthetic data [45] and developing new algorithms that maintain fairness under differential privacy constraints [46]. Understanding the complex relationship between privacy and fairness in SDGs is critical. For example, Fioretto and colleagues [22] highlight how such research can clarify the challenges of decision-making with sensitive data, guide the design of fairer ML systems, and shed light on the social impacts of differential privacy. Additionally, privacy and fairness concerns often overlap, as seen in the use of student ethnicity data, which is both private and essential for fairness assessments [12]. This underscores the need to explore privacy and fairness simultaneously from a data-centric perspective.

Despite this, few studies directly assess both privacy and fairness in synthetic data [6, 10]. Most research has focused on examining the relationship between fairness and data utility, or privacy and utility. Moreover, many earlier privacy evaluations have relied solely on $\epsilon$ as the primary metric (as $\epsilon$ as a privacy parameter also quantifies the privacy level), without considering a broader set of privacy measures. To address this gap, this paper will employ a comprehensive set of privacy evaluation metrics for a more thorough analysis. At the same time, LLM-based tabular SDGs has recently achieved significant breakthroughs [8, 38], yet no studies have compared LLMs to other synthetic tabular SDGs in terms of privacy and fairness. This paper aims to fill that research gap by conducting such a comparison. Finally, our literature review shows that many SDGs designed to balance privacy and fairness primarily achieve this by applying fairness constraints or causal models during data generation [2, 51]. However, common fairness-enhancing pre-processing methods used in real-world ML, such as Disparate Impact Remover, Suppression, and Reweighing, have limitedly been applied to synthetic data. For instance, Bhanot [6] tested a pre-processing method (specifically, Reweighing) on the synthetic data generated by only one type of SDG to see whether it improves fairness in synthetic data. Bhanot's results indicated that while the pre-processing method was effective, the improvement was marginal. This raises our curiosity about whether a broader range of pre-processing methods would be effective on synthetic data generated by different SDGs, and how significant the effects would be.

## 2.2 Synthetic data generation and evaluation

Synthetic data refers to data generated by specially designed mathematical models or algorithms to address a set of data science tasks [28]. The types of synthetic data include text data, tabular data[1], time series data, multimedia data, such as images, audio and video. Other forms can also include geospatial data and graph data, depending on the application. However, since this paper focuses on educational tabular datasets, the emphasis is placed exclusively on synthetic tabular data. Therefore, the following discussion of synthetic data in this paper specifically refers to synthetic tabular data. In addition to enhancing fairness and privacy as mentioned above, synthetic data can also reduce the cost of collecting and labeling real data [28].

The methods for generating synthetic data can be broadly categorized into two types: statistical methods and deep learning methods [13, 21]. Statistical methods have advantages such as fast speed, low computational resource requirements, and manageable parameters. However, they may not be suitable for handling large or complex datasets [26]. On the other hand, deep learning methods, particularly Generative Adversarial Networks (GANs), are renowned for their efficiency and ability to learn underlying patterns in data [21], and they have shown excellent performance across multiple evaluation dimensions [26, 39]. For these reasons, this paper will focus on deep learning methods. Additionally, LLMs are also a type of deep learning method. However, researchers have traditionally considered LLMs to be more adept at generating textual data, with less exploration into their performance in generating tabular data [19]. Nevertheless, the remarkable performance of LLMs has recently garnered significant interest from both academia and industry, leading to the belief that LLMs might lay the groundwork for achieving Artificial General Intelligence [19]. Consequently, scholars have begun exploring the capabilities of LLMs in handling various tabular data tasks, and it has been shown that they can achieve performance comparable to that of representative GAN-based methods [8, 32, 38]. Considering the above situation, this paper will examine two representative deep learning methods: (1) CTGAN (Conditional Tabular Generative Adversarial Network), which performs well in tabular datasets with complex relationships and is a widely used algorithm for synthetic tabular data generation, and (2) DistilGPT2 under the Generation of Realistic Tabular Data (GReaT) framework, is a lightweight LLM that demonstrates strong performance in handling tabular data with heterogeneous feature types.

---

[1]Tabular data refers to data organized into tables, where information is arranged in rows and columns. seen from Krishnamurthi, S., Lerner, B.S. and Politz, J.G., 2017. Introduction to Tabular Data. Available at: https://papl.cs.brown.edu/2016/intro-tabular-data.html

Due to the utility goal of data synthesis (i.e., generating data that is as similar as possible to real data and has similar ML performance), synthetic data tends to retain the distribution of the original data. Therefore, these synthetic data generation models may be susceptible to privacy leakage [28]. Deep neural network-based methods such as CTGAN and DistilGPT2 may be vulnerable to membership inference attacks [28]. If a membership inference attack is successful, it can determine whether a specific input was included in the training data. This compromises privacy by revealing the participation of particular data points in the training dataset, which can lead to unintended privacy breaches. In this context, using differential privacy (DP) for synthetic data generation has become a promising solution [21, 28]. DP is a rigorous privacy concept proposed by [18] to protect sensitive data in dataset disclosures. When DP is combined with synthetic data, the similarity between data points and their corresponding points in the original data does not imply a privacy breach [28], making it more compliant with relevant data privacy regulations. In the field of deferentially private synthetic data generation, there have already been some notable studies. This paper has selected two representative methods for experiments. The first is Private Aggregation of Teacher Ensembles (PATE)-GAN, a model specifically designed for generating high-quality differentially private synthetic data by combining the strengths of GANs and the PATE framework [29]. However, DP-based synthetic data generation may also have some potential drawbacks, such as being overly conservative in privacy settings, which can compromise data utility and reduce the quality of the synthetic data [21]. Therefore, this paper also employs another privacy-preserving SDG, Anonymization Through Data Synthesis Using GAN (ADS-GAN) [57], which aims to generate synthetic data with similar statistical properties to real data while simultaneously reducing the individual identifiability.

Furthermore, synthetic data that reflects the key statistical characteristics of real data also inherits the biases present in data pre-processing, collection, and algorithms [21]. Moreover, minority groups tend to be underrepresented in synthetic data, and the use of DP may exacerbate fairness issues in the original data [21]. Given that this paper aims to explore the relationship between privacy and fairness in SDGs, we will also use a method dedicated to generating fairer synthetic data named DEbiasing CAusal Fairness (DECAF) [51]. This method is chosen for its ability to address fairness concerns by leveraging causal inference techniques, ensuring that the generated synthetic data minimizes bias while maintaining the underlying data distribution [51].

## 3  Methods

This section is organized as follows: First, we briefly describe our datasets, thereafter, we discuss techniques and evaluation metrics. Finally, we provide the experimental details.

### 3.1  Data

The datasets for our experiments, labeled as A, B, and C, are briefly described as follows:

A) Student Math performance dataset from UCI [11]. It contains data related to students' demographics, family history, access to IT facilities, and their Mathematics achievement (i.e., pass or fail) in a Portuguese secondary school. This dataset consists of 395 rows and 30 features. We used 27 features after cleaning the data. We use *sex* as the sensitive attribute for our fairness analysis.

B) Open University Learning Analytics dataset (OULAD) [36]. We randomly took 30% of this dataset that is focused on pass or fail in final results for our experiments. In all, we used records of 5,550 unique students across seven distinct features. We use *disability* as the sensitive attribute for fairness analysis.

C) The law dataset is the longitudinal bar passage data collected from the class that started law school in the fall of 1991, provided by The Law School Admission Council (LSAC) National Longitudinal Study [54]. We use the processed version of the dataset that was used in the fairness study by Kusner and colleagues [35]. This dataset contains six features and 21,792 students records. The target label of this data first year bar exams passage or failure. We use *race* as the sensitive attribute for our fairness analysis.

## 3.2 Techniques and Metrics

In this section we discuss the various SDGs, ML algorithms, fairness algorithms and the respective metrics for the evaluation of privacy and fairness.

*3.2.1 SDG Techniques.* As mentioned in subsection 2.2, this paper utilizes five different SDGs. We will briefly introduce them in this paragraph. (1) **CTGAN** is a GAN-based method designed to model tabular distributions and generate row samples from those distributions. Its unique advantages include the ability to mitigate class imbalances in training data through conditional generators and sampled training [55]. Additionally, CTGAN has shown superior performance in tabular data applications compared to most Bayesian and deep learning methods [55]. (2) **DistilGPT2 (DGPT)** is a Large Language model, which is pre-trained with the smallest version of GPT2. DGPT is trained on OpenAI's WebText dataset [48]. (3) **ADSGAN** introduces a regularization term during training to reduce overfitting to real data, thus lowering the risk of privacy attacks such as membership inference. By balancing this regularization with data utility, ADSGAN achieves both privacy protection and high-quality data generation, making it suitable for scenarios where sensitive data is involved [57]. (4) **PATEGAN** combines the PATE framework with GANs to generate high-quality synthetic data with differential privacy guarantees. It uses an ensemble of teacher models to provide private labels for training the GAN, ensuring privacy without significant degradation of utility [29]. (5) **DECAF** is a GAN-based model specifically designed to generate fair synthetic data for tabular datasets. It incorporates fairness constraints directly into the training process by adjusting the loss function to penalize biased outcomes [51]. This ensures that the generated data treats all demographic groups equitably.

*3.2.2 Privacy evaluation metrics.* As for the privacy evaluation metrics, we used two types of privacy evaluation to provide comprehensive privacy evaluation. The detailed definition of these metrics can be seen in the supplementary file located at this Link.

The first type of privacy evaluation is distance and similarity metrics. Distance and similarity metrics in privacy partially overlap with similarity evaluation. The reasoning is straightforward: if synthetic data is too similar to the original data, there is a high risk to privacy. For distance and similarity metrics, we used Average Jensen-Shannon Distance (JSD) and Wasserstein Distance (WD). The JSD excels at measuring the similarity between categorical synthetic data and real data, while the WD is better suited for measuring the similarity between continuous synthetic data and real data [60]. For all these reasons, JSD and WD were selected for this paper. JSD value is between [0-1], and both larger JSD and WD indicate that the datasets are more dissimilar.

The second type privacy evaluation we use is re-identification risk assessment. It refers to evaluating the risk of real data leakage through re-identification using SDG [26]. Specifically, we used membership inference attack (MIA) and k-anonymization. MIA in a synthetic data environment means that an attacker attempts to identify whether real records were used to train the SDGs [26]. This method can test the robustness of synthetic datasets against adversarial attacks.

For which MIA method we use, we used Block Neural Autoregressive Flow (BNAF) from the DOMIAS framework as an effective density estimator for modeling complex data distributions [33]. This approach enhances the detection of overfitting in generative models, which is key to improving the accuracy of MIA [33]. A higher accuracy is generally a negative outcome, which means the synthetic data is more vulnerable to privacy leakage. K-anonymization, which is a privacy protection measure that ensures an individual's information is indistinguishable from at least k-1 other individuals [47]. A higher score represents better privacy.

*3.2.3 ML and Fairness Algorithms.* For our fairness analysis, we considered 4 **baseline (BL)** ML algorithms, i.e., off-the-shelf ML algorithms *without* any fairness constraints, and 4 pre-processing fairness algorithms. We considered only the pre-processing category of fairness algorithms because it allows us to easily debias the real and synthetic datasets and apply them to downstream ML models. The BL algorithms that we used in this study, namely Random Forest (RF), Logistic Regression (LR), Gaussian Naive Bayes (GNB), and eXtreme Gradient Boosting (XGB), have been extensively used for many classification tasks in LA research [25]. We briefly describe the fairness algorithms we considered as follows. **(1) Suppression (SUP)** [30]: This technique aims to achieve fairness by explicitly excluding the sensitive (demographic) attributes from the data during model training. While some studies such as [43] have criticized this technique that it does not remove unfairness because the sensitive attributes can be correlated and redundantly encoded in other non-sensitive attributes, there are other studies such as [14, 52] that have found that SUP can improve fairness. **(2) Correlation Remover (CoR)** [53]: This technique is part of the fairlearn suite[2] from Microsoft Research. CoR improves upon the limitation of the SUP approach by removing both the sensitive attributes *and* the correlations between the sensitive attributes and the non-sensitive attributes in the training data through the application of linear transformation. **(3) Disparate Impact Remover (DIR)** [20]: This technique removes unfairness by modifying the input features to ensure that their distributions are more similar across the protected and unprotected groups, thereby reducing unfairness in the model's outcomes. **(4) Reweighing (RW)** [30]: This technique achieves fairness by assigning different weights to instances in the dataset based on their group membership and the target label, so that the model trained on this reweighed data treats different groups more fairly, reducing unfairness in the outcomes.

*3.2.4 Fairness evaluation metrics.* To evaluate the fairness of models, we employ 3 popularly used metrics namely ABROCA [23], Error Rate Difference (ERD) [5], and True Positive Rate Difference (TPRD) *aka* equal opportunity [24] and briefly describe them as follows. Firstly, the ABROCA metric measures fairness by calculating the total difference between ROC curves for privileged and unprivileged groups[3] over all decision thresholds. This metric is particularly valuable in LA because it highlights disparities in model predictions across the full range of possible outcomes, providing a complete and threshold-independent view of fairness in predictions. Typically, an ABROCA score ranges from 0 (ideal) to 1 (worst). Mathematically, ABROCA is computed as follows: $ABROCA = \int_{0}^{1} |dROC_b(t) - ROC_c(t)| \, dt$. Where: $ROC_b(t)$ is the ROC curve for the baseline (privileged) group $b$, $ROC_c(t)$ is the ROC curve for the comparison (unprivileged) group $c$, and $t$ is the decision threshold, ranging from 0 to 1. Secondly, the ERD metric measures the disparity in misclassification rates between an unprivileged group $c$ and a privileged group $b$. It helps assess fairness by highlighting which group faces more errors. Since accuracy is the complement of error rate, accuracy difference is the negative of ERD. This also implies that lower ERD scores will lead to lower *accuracy difference* scores. ERD close to 1 means the unprivileged group has significantly more errors than the privileged group, 0 implies equal error rates across both groups (which is ideal), and close to -1 means the privileged group has significantly more errors than the unprivileged group. ERD is calculated as follows: $ERD = Error\ Rate_c - Error\ Rate_b$. Thirdly, the TPRD metric checks if different groups have the same chance of being correctly identified for positive outcomes. TPRD is represented mathematically as: $TPRD = TPR_b - TPR_c$. Where: $TPR_b$ is the True Positive Rate for the privileged group $b$, and $TPR_c$ is the True Positive Rate for the unprivileged group

---

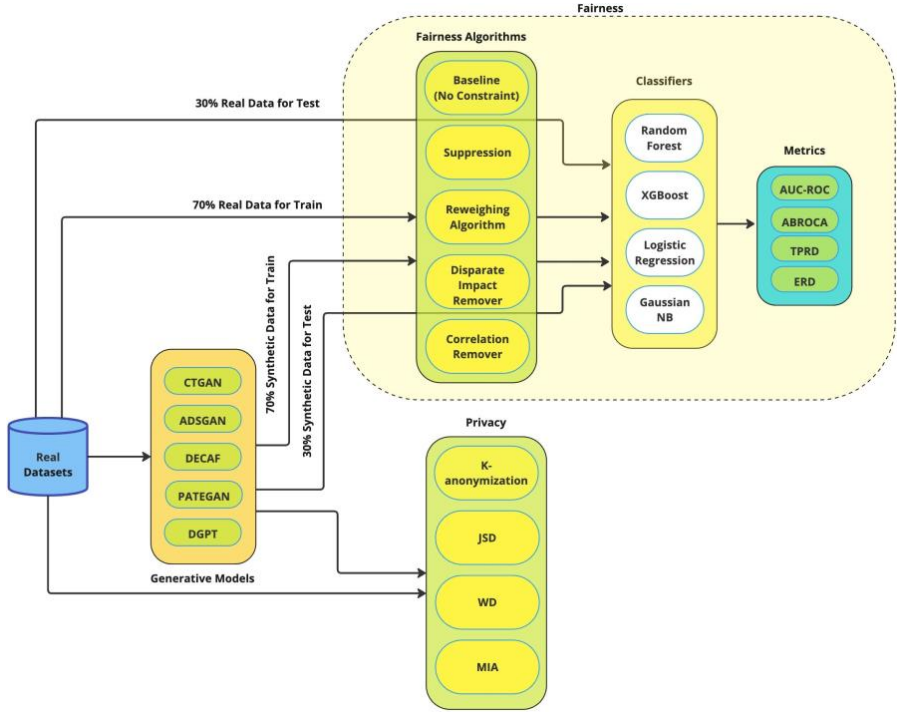

Fig. 1. The overall flow of our experiments. It starts with (1) synthetic data generation and privacy evaluation, (2) training of baseline and fair models on both real and synthetic data, and (3) evaluation of baseline and fair models for fairness and predictive accuracy.

$c$. A TPR value of 0 indicates perfect fairness, a positive value means the unprivileged group has a higher true positive rate, and a negative value means the privileged group has a higher true positive rate.

In this study, we use the *absolute* values for all fairness metrics since our focus is on measuring the *presence of unfairness* and not necessarily determining against which demographic group the models are unfair, as is commonly done in LA research. Furthermore, to make our fairness scores more *"intuitive"* as it were, we normalized the fairness scores as follows: *New Fairness Score* $= 1 − |Old Fairness Score|$. This way, fairness scores closer to 1 are best and those closer to 0 are worse. While our main focus is to investigate the fairness of the models with respect to these 3 metrics, is is also important to investigate if the models are making accurate decisions to begin with. For that reason, we also evaluate the predictive accuracy of the models with respect to the widely used Area Under the ROC curve (AUC-ROC) metric [23, 50].

### 3.3 Experiments

Recall our research objectives are (1) to determine which synthetic data generator performs best in terms of balancing both privacy and fairness and, (2) to investigate that if the fairness of synthetic data is less ideal, how can we improve the fairness of synthetic datasets with pre-processing fairness algorithms? As shown in Figure 1, we first generate synthetic data using five SDGs and then evaluate its privacy and fairness to answer RQ1. Subsequently, we apply fairness algorithms to improve the fairness of the synthetic data, and then evaluate it again using the same fairness metrics to answer RQ2. Throughout the experiment, we used the Synthcity Python library [47] to generate data and conduct privacy evaluations. The entire experiment was conducted in Google Colab, utilizing an Nvidia A100 GPU to enhance computational speed.

*3.3.1 Privacy and Fairness Analysis.* To answer the RQ1, we have used 5 SDGs. We adhere to its recommended default parameters in Synthcity. Specifically, the epsilon value for PATEGAN is 1. For the privacy evaluation of the synthetic data, we used JSD and WD to assess whether the synthetic data is overly similar to the real data, which could lead to privacy leakage. We also employed MIA and k-anonymity for additional evaluation. The privacy evaluation was repeated twice, and average values were taken to minimize the effect of randomness.

After evaluating the synthetic datasets on the privacy metrics, we followed up by training the baseline models on the real and synthetic datasets and evaluated them using the fairness metrics. Before we discuss the experimental details pertaining to fairness in the next paragraphs, we briefly give a preamble regarding the train-test split criterion of the datasets. For all the fairness-related experiments, the train-test split of our datasets follow two paradigms. For the first paradigm called *Same Train, Real Test*, we split both the real and the synthetic datasets using the 70/30 split. We train *all* the models on 70% of each of the datasets—both real and all synthetic datasets. However, during testing, we test *all* the models on the 30% test set of the real data *only*. By doing so, it allows us to hold all the models to the same standard, ensuring *objective and fair* comparison. Our primary focus in this paper, and consequently, our results and discussion will *only* be on the first paradigm. The second paradigm called *Same Train, Same Test* is similar to to the *Same Train, Real Test*, however, instead of testing all the models on *only* 30% real test data, each model is tested on their respective 30% test data as per the training data. This allows for us to measure the fairness of the models when the test dataset comes from the same distribution as the training data. The second paradigm is for demonstration purpose only. Hence, we present the results for the second paradigm in the Supplementary results which can be found here here.

Moving on to the experimental details. We trained all the four baseline models (i.e., RF, XGB, LR, and GNB) on 70% of each of the datasets. For each model, we identified the the optimal hyper-parameters by performing an exhaustive grid search over a specified search space and evaluated the optimal hyper-parameters by 5-fold stratified cross-validation. We then evaluated the fairness of the predictions of each model on the 30% test dataset in terms of ABROCA, ERD, and TPRD. We also evaluated the predictive accuracy of the models in terms of AUC-ROC. Note that we did the training and testing of all models for both paradigms—*Same Train, Real Test* and *Same Train, Same Test*.

To help us identify which SDG is able to strike the best balance between privacy and fairness, we finally compared the various models in terms of privacy *vs.* fairness.

*3.3.2 Fairness Improvement Analysis.* To investigate the RQ2, we applied each of the 4 pre-processing techniques (i.e., SUP, CoR, DIR, and RW) to *debias* each of the 70% training datasets. Some debiasing techniques require certain specifications to determine the degree of fairness required. For the CoR approach, we set the $\alpha = 1.0$ to ensure maximum filtering of biased information. Similarly, for the DIR, we set the *repair level* = 1.0 to ensure maximum fairness. After debiasing all the training datasets, we performed all experiments *again* in the same manner as we did in the RQ1. Everything including model training and evaluation remains the same. The only thing that changes is that this time around, the training data is *debiased*. We analyzed the improvement in fairness or lack thereof of all the synthetic datasets after applying the pre-processing algorithms to them. We present the results of the experiments in Section 4.

# 4 Results

We report selected results here while the additional results are in the *double-blind* supplementary file which can be accessed by Link. Nonetheless, the results presented here extensively capture all our findings.

## 4.1 Privacy *vs.* Fairness Among SDGs

This section aims to answer the RQ1, which SDGs strikes the best balance between privacy and fairness. The results are presented in Table 1 and Figure 2. Table 1 contains the *overall fairness* (hereinafter called fairness) *vs.* the *overall privacy*

(hereinafter called privacy) results across all the 3 datasets. We operationalize fairness as follows:

$$Fairness = \frac{3 - |ABROCA| - |ERD| - |TPRD|}{3}$$

Additionally, let $WD_{max}$, $JSD_{max}$, K-Anonymity$_{max}$, and MIA Accuracy$_{max}$ be the maximum value for each privacy metric among all metric values respectively. We operationalize privacy using the following formula:

$$Privacy = \frac{\frac{WD}{WD_{max}} + \frac{JSD}{JSD_{max}} + \frac{\text{K-Anonymity}}{\text{K-Anonymity}_{max}} + 1 - \frac{\text{MIA Accuracy}}{\text{MIA Accuracy}_{max}}}{4}$$

Figure 2 shows the Parento-Frontier plot displaying which SGDs are the best in terms of privacy and fairness and the *trade-offs* that come with optimizing for either fairness or privacy.

In Table 1 and Figure 2, DECAF shows the best balance between privacy and fairness in the three datasets. In dataset A, DECAF performs well and is at the Pareto frontier. Similarly, in datasets B and C, DECAF continues to perform well, especially in dataset C, showing a strong balance between privacy and fairness. One reason may be that as a fairness-orientated algorithm, it supports debasing during the inference stage by removing specific causal paths, effectively reducing unfair factors and meeting user-defined fairness requirements [51]. The reason for good privacy performance is that to remove bias, DECAF allows the deletion of some protected attributes (such as age, and gender), which causes a slight decrease in similarity with the original dataset, thereby improving privacy.[51].

For utility-focused SDG, both DGPT and CTGAN tend to focus on fairness and privacy respectively, but they show different degrees of imbalance. DGPT leans more towards fairness compare to CTGAN, the evidence is the high fairness scores on datasets A and B in Table 1. In terms of balancing, DGPT is imbalanced on dataset A with the lowest privacy score, more balanced on dataset B ranking second in both privacy and fairness, and performs moderately on dataset C, ranking second-to-last. While CTGAN exhibits extreme imbalance, with significant disparity between fairness and privacy scores on dataset A, the lowest scores for both on dataset B, and the lowest fairness but second-highest privacy score on dataset C. The relative imbalance of CTGAN and DGPT in privacy and fairness may stem from their primary focus on data utility—ensuring that synthetic data closely resembles real data in structure and ML performance [55] [8]. Their goals differ from those of privacy-focused SDGs like ADSGAN and PATEGAN, or fairness-oriented SDGs like DECAF. This observation further supports previous research that demonstrated the trade-off relationship between privacy, fairness, and utility [40].

When comparing privacy-focused ADSGAN and PATEGAN with other SDG, both tend to emphasize privacy more than models like DGPT and CTGAN, but to varying degrees of balance. Overall, ADSGAN emphasizes fairness more than PATEGAN, with a high fairness score of 0.95 but a low privacy score of 0.62 on dataset A. However, it is more imbalanced on datasets B and C, with privacy scores of 0.11 (second-to-last) and 0.44 (lowest), prioritizing fairness at the expense of privacy. The performance of PATEGAN is even more extreme. On dataset A, it shas the lowest fairness score but the highest privacy score. On dataset B, PATEGAN is relatively balanced, with the highest privacy score, but not the worst fairness score. On dataset C, PATEGAN's performance is unstable and even shows the worst privacy score. Overall, ADSGAN provides a more balanced approach across the dataset, while PATEGAN is more unbalanced, prioritizing privacy over fairness to a greater extent. The results are consistent with expectations, as PATEGAN is an SDG that applies DP and has a relatively conservative epsilon value, which is expected to exhibit more extreme trade-offs compared to ADSGAN.

Table 1.  Fairness and Privacy for all Datasets. **Boldened** scores are the highest and those in <span style="color:red">red</span> are the lowest.

| Data | Dataset A | | Dataset B | | Dataset C | |
|---|---|---|---|---|---|---|
| | Fairness | Privacy | Fairness | Privacy | Fairness | Privacy |
| Real Data | 0.93 | - | 0.86 | - | 0.87 | - |
| ADSGAN | **0.95** | 0.62 | 0.84 | 0.11 | 0.85 | 0.44 |
| CTGAN | 0.93 | 0.62 | 0.83 | 0.08 | 0.75 | 0.50 |
| DECAF | 0.91 | 0.69 | **0.91** | 0.13 | **0.91** | **0.64** |
| DGPT | 0.94 | 0.60 | 0.86 | 0.17 | 0.82 | 0.46 |
| PATEGAN | 0.90 | **0.71** | 0.84 | **0.33** | 0.90 | 0.44 |

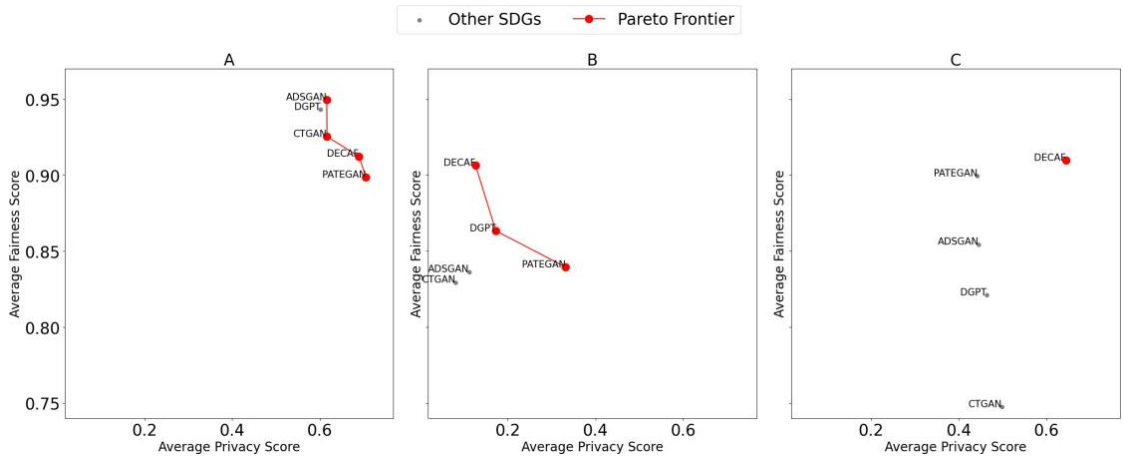

Fig. 2.  Pareto frontier illustrating the trade-off between fairness and privacy. Red-labeled points represent optimal solutions, balancing both objectives, while gray points are suboptimal. The frontier highlights the trade-offs between improving fairness and preserving privacy.

## 4.2   Application of Fairness Algorithms to Improve Fairness of Synthetic Datasets

We observed that the application of fairness algorithms can improve the fairness of the models trained on the synthetic datasets. For instance, consider Table 2. This table depicts the *overall* percentage of improvement (*+ve*) or exacerbation (*-ve*) of fairness of *all* models across *all* fairness metrics for the various datasets after the pre-processing algorithms have been applied to the datasets. We observed that the SDGs tend to enjoy significant improvement in fairness after the fairness algorithms were applied. Noteworthy among the SDGs is the CTGAN. In fact, on the Dataset C, fairness improved for the CTGAN by 21.5% relative to the baseline. Interestingly, we observed that the CTGAN tends to perform well when paired up with the RW and the SUP fairness techniques. For example, for both datasets B and C, the CTGAN-RW and CTGAN-SUP combinations had the most improvement in fairness relative to their respective baselines. The Figure 3 reinforces these findings, showcasing upward and downward trends in fairness across datasets when the fairness algorithms are employed. Notably, CTGAN shows stable and significant improvements across datasets B and C when RW and SUP are used, indicating that these algorithms may be particularly effective when combined with CTGAN.

However, it should be noted that the effectiveness of different fairness algorithms on the synthetic datasets is not consistent. For example, CTGAN performs remarkably well on SUP across all datasets, showing an impressive 17.7% improvement in Dataset C. In contrast, ADSGAN exhibits largest negative value (-3.8%) on CoR in Dataset A. Overall, from the results of the three experimental datasets, it is difficult to conclusively state which fairness algorithm is most beneficial for improving synthetic data. It can only be observed that certain fairness algorithms are better suited to synthetic data generated by specific SDG.

It is evident that fairness algorithms tend to enhance fairness more effectively on synthetic datasets compared to real datasets. Across all datasets, the real data rarely exhibits the highest fairness improvement after applying fairness algorithms. For instance, the only instance where real data performed best was with RW on Dataset A (4% improvement), yet this improvement was marginally higher than the corresponding improvement for the DPGT synthetic dataset (3.1%). Even when fairness for real data improved by 11.5% on Dataset B (using RW), the DECAF synthetic dataset outperformed it with a 17.7% improvement. This observation suggests that while fairness algorithms do enhance real data, their impact is not as pronounced as on synthetic data, which shows consistently higher improvements. A key insight from these results is that the SDG itself may inherently address some fairness concerns, functioning similarly to a fairness algorithm. This aligns with existing literature which shows that SDGs can enhance algorithmic fairness [1]. Therefore, the combination of SDG with pre-processing fairness algorithms appears to be a promising approach to achieving fairer models. Moreover, the significant improvements observed across synthetic datasets indicate that SDGs, when combined with fairness algorithms, may offer a more robust and effective method for mitigating bias than fairness algorithms applied to real data alone. But another point worth noting is that the performance of synthetic data on the smaller data set (sample size = 395) of Dataset A is not very stable. While the overall improvement of fairness methods is greater on synthetic data than on real data, some fairness algorithms applied to synthetic data produce negative results. This may be due to synthetic data being less stable in smaller datasets compared to larger ones [28].

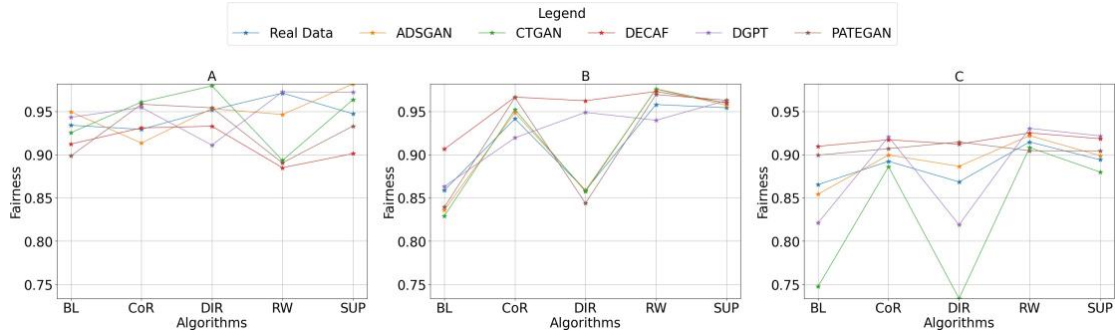

Fig. 3. The trend of fairness improvement or decline with respect to the baseline (BL) after the application of the 4 pre-processing fairness algorithms (i.e., CoR, DIR, RW, and SUP) for both real and synthetic datasets across all models. (Higher fairness is beBer) : $Fairness = \frac{3 - |ABROCA| - |ERD| - |T PRD|}{3}$

Table 2. Percentage Change in Average Fairness Results Compared to the Baseline Across All Models and Datasets

| Data | Dataset A | | | | Dataset B | | | | Dataset C | | | |
|---|---|---|---|---|---|---|---|---|---|---|---|---|
| | CoR | DIR | RW | SUP | CoR | DIR | RW | SUP | CoR | DIR | RW | SUP |
| Real Data | -0.5% | 1.9% | **4.0%** | 1.4% | 9.6% | 0.0% | 11.5% | 11.1% | 3.1% | 0.4% | 5.7% | 3.3% |
| ADSGAN | -3.8% | 0.4% | -0.3% | 3.4% | 13.4% | 2.7% | 16.7% | 14.5% | 5.3% | **3.8%** | 7.9% | 5.2% |
| CTGAN | 3.8% | 5.9% | -3.5% | **4.1%** | 14.8% | 3.4% | **17.7%** | **15.7%** | **18.6%** | -1.8% | **21.5%** | **17.7%** |
| DECAF | 2.1% | 2.3% | -3.0% | -1.2% | 6.6% | 6.2% | 7.3% | 5.9% | 0.8% | 0.3% | 1.7% | 1.0% |
| DGPT | 1.2% | -3.4% | 3.1% | 3.1% | 6.5% | **9.9%** | 8.9% | 11.5% | 12.1% | -0.3% | 13.3% | 12.3% |
| PATEGAN | **6.7%** | **6.2%** | -0.9% | 3.8% | **15.1%** | 0.5% | 15.5% | 14.7% | 0.8% | 1.7% | 0.6% | 0.6% |

Table 3. Average AUC-ROC for Dataset A with the corresponding standard error of the mean (SEM). Best value in **bold** and worst value in <span style="color:red">red</span>

| Data | BL | CoR | DIR | RW | SUP |
|------|-----|-----|-----|-----|-----|
| Real Data | **0.97 ± 0.02** | **0.94 ± 0.02** | **0.94 ± 0.02** | **0.93 ± 0.02** | 0.92 ± 0.03 |
| ADSGAN | 0.95 ± 0.02 | 0.92 ± 0.02 | 0.90 ± 0.02 | 0.92 ± 0.02 | **0.93 ± 0.04** |
| CTGAN | 0.92 ± 0.02 | 0.90 ± 0.01 | 0.91 ± 0.02 | 0.86 ± 0.03 | 0.89 ± 0.03 |
| DECAF | <span style="color:red">0.47 ± 0.06</span> | <span style="color:red">0.42 ± 0.06</span> | <span style="color:red">0.48 ± 0.04</span> | <span style="color:red">0.41 ± 0.06</span> | <span style="color:red">0.44 ± 0.02</span> |
| DGPT | 0.94 ± 0.03 | **0.94 ± 0.02** | 0.89 ± 0.04 | 0.90 ± 0.03 | 0.91 ± 0.03 |
| PATEGAN | 0.77 ± 0.02 | 0.75 ± 0.02 | 0.71 ± 0.03 | 0.65 ± 0.02 | 0.71 ± 0.04 |

Table 4. Average AUC-ROC for Dataset B with the corresponding standard error of the mean (SEM). Best value in **bold** and worst value in <span style="color:red">red</span>

| Data | BL | CoR | DIR | RW | SUP |
|------|-----|-----|-----|-----|-----|
| Real Data | 0.62 ± 0.0 | **0.63 ± 0.01** | **0.61 ± 0.01** | **0.60 ± 0.0** | **0.63 ± 0.0** |
| ADSGAN | **0.63 ± 0.0** | **0.63 ± 0.0** | 0.59 ± 0.01 | **0.60 ± 0.01** | 0.62 ± 0.01 |
| CTGAN | 0.62 ± 0.0 | **0.63 ± 0.01** | **0.61 ± 0.01** | 0.59 ± 0.0 | **0.63 ± 0.0** |
| DECAF | <span style="color:red">0.51 ± 0.0</span> | 0.56 ± 0.01 | 0.55 ± 0.01 | 0.52 ± 0.0 | 0.54 ± 0.02 |
| DGPT | 0.53 ± 0.0 | <span style="color:red">0.50 ± 0.01</span> | 0.53 ± 0.01 | 0.51 ± 0.01 | 0.54 ± 0.02 |
| PATEGAN | 0.52 ± 0.01 | 0.51 ± 0.01 | <span style="color:red">0.50 ± 0.01</span> | <span style="color:red">0.48 ± 0.01</span> | <span style="color:red">0.46 ± 0.02</span> |

Table 5. Average AUC-ROC Results for Dataset C with the corresponding standard error of the mean (SEM). Best value in **bold** and worst value in <span style="color:red">red</span>

| Data | BL | CoR | DIR | RW | SUP |
|------|-----|-----|-----|-----|-----|
| Real Data | **0.84 ± 0.01** | **0.82 ± 0.0** | **0.83 ± 0.0** | **0.82 ± 0.01** | **0.84 ± 0.01** |
| ADSGAN | 0.81 ± 0.01 | 0.81 ± 0.01 | 0.79 ± 0.02 | 0.80 ± 0.01 | 0.83 ± 0.01 |
| CTGAN | 0.81 ± 0.01 | 0.81 ± 0.01 | 0.80 ± 0.01 | 0.80 ± 0.02 | **0.84 ± 0.01** |
| DECAF | 0.71 ± 0.04 | 0.72 ± 0.01 | 0.74 ± 0.02 | 0.73 ± 0.02 | 0.73 ± 0.02 |
| DGPT | 0.77 ± 0.04 | 0.77 ± 0.04 | 0.78 ± 0.02 | 0.76 ± 0.06 | 0.78 ± 0.05 |
| PATEGAN | <span style="color:red">0.57 ± 0.04</span> | <span style="color:red">0.48 ± 0.08</span> | <span style="color:red">0.53 ± 0.07</span> | <span style="color:red">0.53 ± 0.08</span> | <span style="color:red">0.48 ± 0.1</span> |

## 5 Discussion and Conclusion

Here, we discuss the utility of our findings and provide some implications for practice. First, for RQ1, we found that DECAF strikes the best balance between privacy and fairness. On the other hand, CTGAN and DGPT lean differently toward privacy and fairness, with DGPT achieving a slightly better balance than CTGAN. PATEGAN and ADSGAN are more privacy-focused, with ADSGAN achieving a better balance compared to PATEGAN. The observation that PATEGAN and ADSGAN prioritize privacy aligns with previous research [29, 57]. This may be because PATEGAN applies DP, which has been shown to negatively impact fairness [10]. ADSGAN, although not using DP, employs a GAN-based structure that minimizes identifiability to privacy-preserving. This structure of ADSGAN likely affects key details relevant to downstream fairness, resulting in its privacy-leaning nature [7]. These findings provide valuable guidance for LA practitioners when they use SDG for fairness enhancement or privacy preservation. When privacy and fairness need to be balanced, DECAF may be the optimal choice. In contrast, ADSGAN is better suited for scenarios where privacy is the primary concern, and DGPT is ideal for situations where fairness is emphasized. Both ADSGAN and DGPT while maintaining a relatively good balance. This result implies that when LA practitioners use SDGs, they have to clarify their goals in three dimensions when formulating data strategies but also emphasize that algorithm selection should be based on the specific needs of the application scenario. This helps guide the development of personalized solutions in the LA field, thereby improving the practical usability and trustworthiness of the system.

However, the best balance achieved by DECAF in terms of privacy and fairness is accompanied by its worst performance in utility. This result can be seen in Tables 3-5, which show DECAF's performance in terms of accuracy. From previous literature, fairness and utility have been considered to have an inverse relationship [21], privacy and utility are also recognized as having an inverse relationship [28], and fairness and privacy are similarly considered to have an inverse relationship [21]. Although in our experiments, one SDG (i.e., DECAF) achieved the best balance in the inverse relationship between fairness and privacy, when placed in the broader triangular relationship of fairness, privacy, and utility, it still struggles to maintain a good balance. This indicates that there is still a lack of SDGs capable of balancing fairness, privacy, and utility simultaneously.

For RQ2, we found that applying pre-processing fairness algorithms after generating synthetic data can improve the fairness of the synthetic data. However, different fairness algorithms show inconsistent effectiveness on synthetic data. This finding has implications for generating fair synthetic data, as previous literature has focused on incorporating fairness constraints directly during the generation of synthetic data [46, 51] to achieve fair synthetic data. Our approach demonstrates that using pre-processing fairness algorithms after the synthetic data is generated is also effective in improving fairness. Additionally, another interesting finding is that the improvement in fairness for synthetic data is greater than for real data. The reason may be that synthetic data works well as a fairness algorithm overlaid with pre-processing fairness algorithms. This finding contrasts with the previous results from [6], who applied a single pre-processing fairness algorithm to synthetic data generated by HealthGAN. However, this could be due to the fact that they tested only one SDG model and one pre-processing fairness algorithm. As mentioned earlier, different fairness algorithms show inconsistent performance on synthetic data. The reweighting method they used might not have been well-suited for HealthGAN. This finding not only provides a pathway for improving current fair synthetic data generation research but also points to future directions for fairness research. By optimizing the combination of pre-processing fairness algorithms and synthetic data generation algorithms, fairness can be better ensured, thus laying a foundation for the development of more fair and transparent AI systems.

The limitations of this study are as follows: First, to reduce the impact of randomness and bias, it is generally advisable to generate synthetic data multiple times and average the evaluation metrics over iterations, especially for DP methods where noise addition introduces variability. Due to computational constraints, this approach was not applied in the current study, but we aim to enhance robustness in future work. Second, while many other privacy- and fairness-oriented SDGs exist and perform well in both domains, this study focused on a select group of representative SDGs to ensure a clear and concise analysis. Lastly, this study is limited to tabular data, even though learning analytics encompasses diverse data types such as time series and images. Future research could extend the scope to include these other data types.

## Acknowledgments

This work was funded by...................